\documentclass[11pt]{article}
\usepackage{acl2016}
\usepackage{times}
\usepackage{url}
\usepackage{latexsym}
\usepackage{graphicx}
\usepackage{textcomp}
\usepackage{microtype}
\usepackage{amssymb}

\aclfinalcopy % Uncomment this line for the final submission
\def\aclpaperid{10} %  Enter the acl Paper ID here

\title{Using Distributed Representations to Disambiguate Biomedical and Clinical Concepts}

\author{St\'{e}phan Tulkens$^\bigstar$ \and Simon \v{S}uster$^{\bigstar\spadesuit}$ \and Walter Daelemans$^\bigstar$ \\
\begin{tabular}{c c}
$^\bigstar$CLiPS & $^\spadesuit$CLCG \\ 
University of Antwerp & University of Groningen\\ 
Prinsstraat 13 & Oude Kijk in 't Jatstr. 26 \\ 
2000 Antwerpen & 9700AS Groningen \\ 
Belgium & The Netherlands\\
\{firstname.lastname\}@uantwerpen.be & \\
\end{tabular}
}
\date{}

\begin{document}
\maketitle
\begin{abstract}
In this paper, we report a knowledge-based method for Word Sense Disambiguation in the domains of biomedical and clinical text. We combine word representations created on large corpora with a small number of definitions from the UMLS to create concept representations, which we then compare to representations of the context of ambiguous terms. Using no relational information, we obtain comparable performance to previous approaches on the MSH-WSD dataset, which is a well-known dataset in the biomedical domain. Additionally, our method is fast and easy to set up and extend to other domains. Supplementary materials, including source code, can be found at \url{https://github.com/clips/yarn}
\end{abstract}

\section{Introduction}

Word Sense Disambiguation (WSD) is a procedure in which an ambiguous term or concept is assigned a single sense appropriate for that context, and is an important step in the creation of a semantic representation of a document \cite{ide1998introduction}. While performing WSD will benefit most natural language processing applications, disambiguation of concepts is a critical component of applications operating on clinical and biomedical text, in which the same word can denote differing concepts, and may thus elicit radically different responses.

Compounding this problem of ambiguity is the fact that clinical text, in general, is noisier than other domains, and contains a large variety of abbreviations, some of which may be specific to a single hospital or physician. Additionally, there is a marked absence of large volumes of annotated clinical text, even for English, which presents a problem for supervised approaches to Word Sense Disambiguation. For other languages, such as Dutch, there exist no freely available annotated corpora of clinical text.

A first step towards solving this problem could be the use of distributed representations. Where a more traditional word representation, such as a TF-IDF bag-of-words (BoW) representation, carries frequency information, distributed representations encode semantic information. A big advantage to using these representations is that they can be generated from large corpora of unlabeled text, and can be trained on very large corpora in a reasonable amount of time. These representations, especially when trained using neural architectures such as \texttt{word2vec} \cite{mikolov2013}, have been shown to improve performance on a variety of tasks when compared to more traditional BoW representations.

We hypothesize that these kinds of distributional representations are well-suited for WSD in the clinical and biomedical domain because of the lack of training data, and the large terminological variety. We present a knowledge-based approach to Word Sense Disambiguation which creates concept representations by combining definitions from the
Unified Medical Language System (UMLS) with distributed representations. We test our hypothesis on the MSH-WSD, which is a well-known dataset for WSD in the biomedical domain.

\section{Related Research}\label{sec:rr}

All knowledge-based methods we review use the Unified Medical Language System\textsuperscript{\textregistered} (UMLS) Metathesaurus\textsuperscript{\textregistered} \cite{umls} as a knowledge base, possibly augmented with external sources, such as MeSH\textsuperscript{\textregistered}-indexed abstracts. Generally speaking, the UMLS contains two separate information sources that are suitable for use in disambiguation: the concept unique identifier (CUI), which is a unique label for each concept, and the semantic type (ST), which is a set of 135 broad labels such as ``Animal'' or ``Chemical''. In general, a word is only considered disambiguated if the correct CUI can be selected; hence, as \newcite{mcinnes2013} note, approaches based on semantic types are not able to disambiguate between approximately 12\% of concepts, as some concepts with the same surface form have an identical ST, but a different CUI. 

In terms of approaches using ST, \newcite{humphreys2005} create one vector for each semantic type by creating a BoW representation of all words that denote that semantic type. For each ambiguous term, a target word vector is created by taking a window of words from the right and left of the term. The concept which is associated with the ST with the lowest cosine distance is then taken to be the correct sense of the term. Similarly, \newcite{alexopoulou} create a method which finds the closest concept based on a combination of co-occurrence with other semantic types and ontological similarity through \emph{is-a} relationships. 

Closest to our approach is the machine readable dictionary (MRD) approach \cite{mcinnes2008,jimenoyepes2011}, which uses definitions from the UMLS to create concept vectors by creating BoW representations of concepts using all definitions of the concept and those of related concepts. This BoW representation contains TF-IDF values where D is the number of concepts in which a word appears, thereby reducing the influence of general words which occur in many concepts. These representations are then compared to the vectorized contexts of the ambiguous terms using cosine distance. A refinement of MRD, called second-order co-occurrence MRD (2-MRD) \cite{mcinnes2008}, replaces each word in a definition by a vector which contains TF-IDF values of co-occurrence counts, thereby associating each word with a context.

\newcite{mcinnes2013} introduce UMLS::SenseRelate, an approach which is based on \newcite{pedersen2004}'s WordNet::SenseRelate. In this system, each possible sense for an ambiguous term is assigned a distance-weighted score based on the \emph{concepts} of the terms surrounding it, where the concepts of the surrounding terms are determined using UMLS::Similarity \cite{mcinnes2009}.

\newcite{jimenoyepes2014} present so-called step models, which calculate the probability of a word occurring with a certain concept by considering the number of times a word occurs in the definitions of that concept and its related concepts. It then steps through the UMLS-defined ontology of concepts, and refines the probabilities for each word and each concept based on the relations within the ontology. 

Finally, \newcite{chen2014unified} present an approach for general WSD which uses word embeddings coupled with WordNet \cite{fellbaum1998wordnet} as a resource to perform sense disambiguation, and which creates sense-specific word embeddings from these sense-disambiguated word representations.

\section{Materials}\label{sec:materials}

\subsection{Test Corpus}\label{sec:corpus}

We use the MSH-WSD corpus \cite{jimenoyepes2011}, which consists of a set of 203 ambiguous terms, each associated with multiple concepts, to evaluate our approach. Of the 203 terms in the corpus, 106 are regular terms, 88 are acronyms, and 9 can be acronyms and regular terms. For each of these concepts, up to 100 MeSH abstracts were retrieved, resulting in a set of 37,888 abstracts. In our approach, all abstracts were pre-processed using the tokenizer from the Pattern package \cite{patternref}, and all stop words were removed using the English stop word list from \texttt{scikit-learn} \cite{sklearn}.

\begin{table}[t!]
\centering
\begin{tabular}{p{1.9cm}| p{1.1cm} p{1.7cm} p{1.2cm}}
\hline
         & \textbf{Medline}  & \textbf{Mimic-III}  & \textbf{Bioasq} \\          \hline
\textbf{Corpus size} & 920,081 & 13,097,844  & -\\
\textbf{Vocabulary} & 196,960 & 71,663 & 1,701,632 \\
\textbf{Dimension} & 320 & 320 & 200 \\
\hline
\end{tabular}
\caption{The number of words in the corpus, the resulting vocabulary size, and the dimension of the resulting vectors.}
\label{corpora}
\end{table}

\subsection{Word vectors}\label{sec:vectors}

We evaluate our approach using three sets of vectors: The first set was trained on a small set of Medline abstracts\footnote{The specific IDs of these abstracts are available in the online appendix.}, and a second set of vectors created on the entirety of the MIMIC-III corpus of clinical notes \cite{mimicref}. For both sets, we used the \texttt{word2vec} implementation from \texttt{gensim} \cite{gensim}, using skipgram with negative sampling, a frequency cutoff of 5 and a negative sampling of 15. 
Additionally, we used a third set of vectors, available from the BioASQ organisers\footnote{Available on the BioASQ website.}, which was trained on a much larger set of Medline abstracts.\footnote{While we concede that the BioASQ corpora might contain abstracts from the MSH dataset, it does not contain any explicit labeled information that might be used in disambiguation.} The model statistics are visualized in Table \ref{corpora}.

\section{Approach}\label{sec:approach}

Similar to the 2-MRD approach detailed above, our approach creates \emph{concept vectors} by replacing each word in every definition by the vector representation of that word. This creates an  $M \times n$ matrix for each definition, where $M$ is the dimensionality of the word vectors, and $n$ the number of words contained in that definition. Following this, for each definition, we then obtain a single vector of dimensionality $M$ by applying a compositional function to the matrix, thereby obtaining so-called \emph{definition vectors}, which represent the entire meaning of the definition in one vector. Each concept can then be represented by a $M \times d$ matrix, where $d$ is the number of definitions that a concept has in the UMLS. Finally, we apply a second composition function to this matrix, thereby obtaining a single vector of dimensionality $M$ which represents the combined meaning of all definitions for that concept, i.e. a \emph{concept vector}.

\begin{figure}[t]
  \centering
    \includegraphics[width=0.4\textwidth]{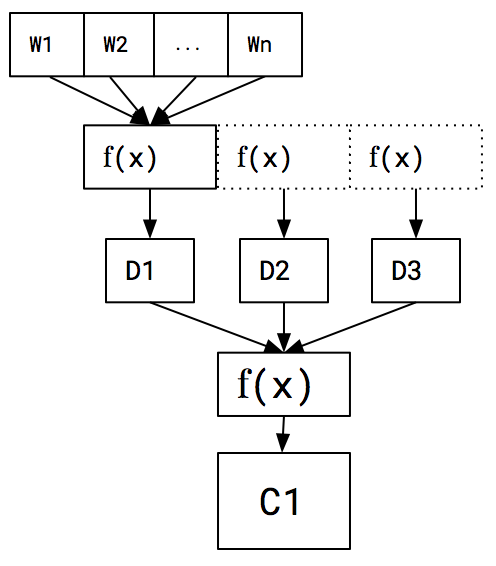}     
\caption{Our model represents a concept by replacing all words $W$ in a definition $D$ by their vectors, and then composing these into a definition vector with a function $f(x)$. For each concept, all definition vectors $D$ are then composed into a concept vector $C$ using a second composition function $g(x)$.}
\label{fig}
\end{figure}

For each abstract in the test corpus, we first locate each ambiguous term through a simple lookup. For each located term in the abstract we create a vector representation by retrieving all words in a window of size $w$ surrounding the ambiguous term, and replacing the words by their vectors. Note that this window does not include the ambiguous term itself. These collections of vectors are then combined into $M$-dimensional vectors using the same composition function as above. This is done separately for each term occurrence within a single document, creating a $M \times x$ matrix, where $x$ is the number of times the ambiguous term occurs in a single document. These are then combined in an $M$-dimensional \emph{term vector} using the same composition we used for the concepts, above. A schematic representation of our model is given in Figure \ref{fig}. 

Because all concept and term vectors are created using the same distributed vectors and compositional functions, the vector space in which they are placed is also comparable. Hence, for each ambiguous word we encounter, we can use the cosine distance between the abstract vector of the ambiguous utterance and each possible sense of that word to determine the correct sense. This makes our approach very similar to the \emph{Lesk} family of approaches \cite{lesk}.

In terms of composition function we experimented with elementwise multiplication, averaging and summation, all of which are unordered compositional functions \cite{mitchell2008}. In addition, it is worth noting that there's still a lively debate whether ordered composition actually leads to better results for estimating document-, or sentence-level meaning, when compared to unordered composition \cite{iyyer2015,socher2013}. 

\section{Results}\label{sec:results}

\begin{table*}[ht!]
\centering
\begin{tabular}{l|p{0.6cm}p{0.6cm}p{0.6cm}p{0.6cm}p{1.3cm}p{1cm}p{1cm}p{1cm}l}
\hline
         & \textbf{med}  & \textbf{mim}  & \textbf{bio}  & \textbf{MRD}  & \textbf{2-MRD} & \textbf{0-step} & \textbf{2-step} & \textbf{r-step} & \textbf{UMLS::SenseRelate} \\
         \hline
\textbf{Accuracy C} & 0.80 & 0.69 & 0.84 & 0.81 & 0.78  & 0.82   & 0.86   & 0.89   & 0.75            \\
\textbf{Accuracy U} & 0.72 & 0.63 & 0.75 & - & -  & - & - & - & -
\end{tabular}
\caption{Results using constrained (C) and unconstrained (U) terms.}
\label{tab:results}
\end{table*}

\begin{table}[ht!]
\centering
\begin{tabular}{r|l}
\hline
\textbf{Term} & \textbf{Accuracy} \\
\hline
\textbf{DE} & 0.31 \\
\textbf{Hemlock} & 0.4 \\
\textbf{Brucella Abortus} & 0.46 \\
\textbf{WT1} & 0.46 \\
\textbf{Murine Sarcoma Virus} & 0.47 
\end{tabular}
\caption{The 5 lowest-performing terms.}
\label{table:discuss}
\end{table}

The accuracy scores obtained by our models using the different word vectors are displayed in Table \ref{tab:results}. $med$, $mim$ and $bio$ denote the vectors created on the small Medline corpus, the Mimic-III corpus and the BioASQ vectors, respectively. We consider both a constrained and an unconstrained version of the task. For each word, the constrained version of the task only considers the senses present in the MSH-WSD dataset as possible targets. The unconstrained version considers all concepts which are denoted by the ambiguous term in the 2015AB version of the UMLS as possible targets. The term \texttt{cortex}, for example, only has 2 concepts associated with it in the MSH-WSD dataset, while in the 2015AB UMLS release it can denote 5 separate concepts. Because the unconstrained version of the task considers all words, it therefore gives a better indication of real-life performance. 

Accuracy C and U denote that the scores were obtained in the constrained settings and unconstrained setting, respectively. All reported scores use a window size of 6, which was optimized on a randomly selected set of 20 terms from the MSH-WSD set. Varying the window size had negligible results: all window sizes over 6 had comparable results, and increasing the window size over 30 causes a (small) decline in results. This is in line with \newcite{mcinnes2013}, who report a positive effect of window size that quickly tapers off for window sizes $> 10$. Concerning the composition functions, summation and averaging as first and second order composition function worked best, while using element-wise multiplication did not work well in any case. Where possible, we display the self-reported scores from the relevant papers on the same dataset.  

A first thing to note is the large difference in accuracy when changing the set of word representations, especially the difference between the Medline vectors and the vectors derived from the Mimic-III corpus. It is currently unclear what causes these performance differences, although it is likely that the small vocabulary, caused by the noisiness of the clinical data in the MIMIC-III corpus, reduces performance. Compared to previous approaches, our approach outperforms the MRD, 2-MRD, and UMLS::SenseRelate approaches, but does not manage to improve on the scores of the step models. Recall, however, that the step models largely rely on relationships in the UMLS ontology to estimate concept relatedness.

To compare how our models improved when including relation information, we also experimented with adding definitions of related concepts, i.e. concepts which had a sibling, parent or child relationship to each concept. In contrast to patterns observed in earlier work, this did not have a significant, and often a detrimental, effect on performance. Note that this makes our model entirely independent of the actual UMLS hierarchy, and more flexible as a result, as we only use the mappings from definition to CUI for disambiguation, and no other information, such as relations or semantic type. In addition, our system is also fast: on a consumer-grade laptop, our approach takes 10 seconds to vectorize and disambiguate all abstracts in the MSH dataset, not taking into account the time it takes to load the embeddings into memory. 

Our approach obtains an accuracy of $> 90\%$ on 103 terms, showing that it is able to disambiguate a large variety of terms. For some terms, however, the performance was below random guessing. These are shown in Table \ref{table:discuss}. The pattern of errors is quite clear: Our approach has trouble with disambiguation if the definitions of the concepts themselves are lexically very similar. As an example, on the term \texttt{Hemlock} our approach performs below chance level because one of the concepts denotes a family of poisonous plants, while the other reports a tree, also called hemlock, the description of which mentions that it is explicitly \emph{not} poisonous. We expect these kinds of problems to be alleviated with the addition of more data. 

\section{Conclusion and future work}\label{sec:discussion}

In this paper we presented a novel approach to WSD in the biomedical domain which achieves comparable performance to existing methods without incorporating relational information from an ontology. This makes the approach easily transferable to other languages, for which such ontologies might not exist, and to other domains. The large variation in accuracy when changing sets of word embeddings also raises interesting prospects for improvement; better word representations will lead to an improvement in our approach without modifying the approach itself. Additionally, we would like to experiment with different composition functions for composing the definition and concept vectors.

\section*{Acknowledgments} 
Part of this research was carried out in the framework of the Accumulate IWT SBO project, funded by the government agency for Innovation by Science and Technology (IWT). We would also like to thank Elyne Scheurwegs for making the small set of Medline abstract available to us.
%\bibliography{acl2016}
\bibliography{acl2016}
\bibliographystyle{acl2016}

\end{document}